\documentclass[twocolumn]{article}

\usepackage{PRIMEarxiv}

\usepackage[utf8]{inputenc} 
\usepackage[T1]{fontenc}    
\usepackage{hyperref}       
\usepackage{url}            
\usepackage{booktabs}       
\usepackage{amsfonts}       
\usepackage{nicefrac}       
\usepackage{microtype}      
\usepackage{lipsum}
\usepackage{fancyhdr}       
\usepackage{graphicx}       
\graphicspath{{media/}}     
\usepackage{xcolor}
\usepackage{colortbl}
\usepackage{subcaption} 
\usepackage{abstract}
\usepackage{xspace}
\usepackage{amsmath}
\usepackage{algorithm}  
\usepackage{algorithmic}  
\usepackage{multirow}
\usepackage{fontawesome}
\usepackage{color}
\usepackage{bm}

\title{TASP: Topology-aware Sequence Parallelism
}

\author{
  Yida Wang$^{\text{1,3}}$\\
  \And
  Ke Hong$^{\text{2,3}}$\\
  \And
  Xiuhong Li$^{\text{3}}$\\
  \AND 
  Yuanchao Xu$^{\text{1}}$~$^{\text{\faEnvelope}}$\\
  \And
  Wenxun Wang$^{\text{2}}$ \\
  \And
  Guohao Dai$^{\text{4,3}}$\\
  \And 
  Yu Wang$^{\text{2}}$~$^{\text{\faEnvelope}}$\\
  \AND
  \large{{\faEnvelope~}\texttt{xuyuanchao@cnu.edu.cn, yu-wang@tsinghua.edu.cn}}
}

\begin{document}

\twocolumn[
  \maketitle
]
\def\thefootnote{$^1$}\footnotetext{Capital Normal University}\def\thefootnote{\arabic{footnote}}
\def\thefootnote{$^2$}\footnotetext{Tsinghua University}\def\thefootnote{\arabic{footnote}}
\def\thefootnote{$^3$}\footnotetext{Infinigence-AI}\def\thefootnote{\arabic{footnote}}
\def\thefootnote{$^4$}\footnotetext{Shanghai Jiao Tong University}\def\thefootnote{\arabic{footnote}}
\def\thefootnote{\faEnvelope}\footnotetext{Corresponding authors: Yuanchao Xu, Yu Wang}\def\thefootnote{\arabic{footnote}}
\begin{abstract}
Long-context large language models (LLMs) face constraints due to the quadratic complexity of the self-attention mechanism.
The mainstream sequence parallelism (SP) method, Ring Attention, attempts to solve this by distributing the query into multiple query chunks across accelerators and enable each Q tensor to access all KV tensors from other accelerators via the Ring AllGather communication primitive. However, it exhibits low communication efficiency, restricting its practical applicability. This inefficiency stems from the mismatch between the Ring AllGather communication primitive it adopts and the AlltoAll topology of modern accelerators. A Ring AllGather primitive is composed of iterations of ring-styled data transfer, which can only utilize a very limited fraction of an AlltoAll topology.

Inspired by the Hamiltonian decomposition of complete directed graphs, we identify that modern accelerator topology can be decomposed into multiple orthogonal ring datapaths which can concurrently transfer data without interference. Based on this, we further observe that the Ring AllGather primitive can also be decomposed into the same number of concurrent ring-styled data transfer at every iteration. Based on these insights illustrated in Figure~\ref{fig:intro}, we propose \textbf{TASP}, a topology-aware SP method for long-context LLMs that fully utilizes the communication capacity of modern accelerators via \textit{topology decomposition} and \textit{primitive decomposition}.
Experimental results on both single-node and multi-node NVIDIA H100 systems and a single-node AMD MI300X system demonstrate that TASP achieves higher communication efficiency than Ring Attention on these modern accelerator topologies and achieves up to 3.58$\times$ speedup than Ring Attention and its variant Zigzag-Ring Attention. The code is available at \url{https://github.com/infinigence/HamiltonAttention}.
\end{abstract}


\begin{figure}
  \centering
  \includegraphics[width=0.99\columnwidth]{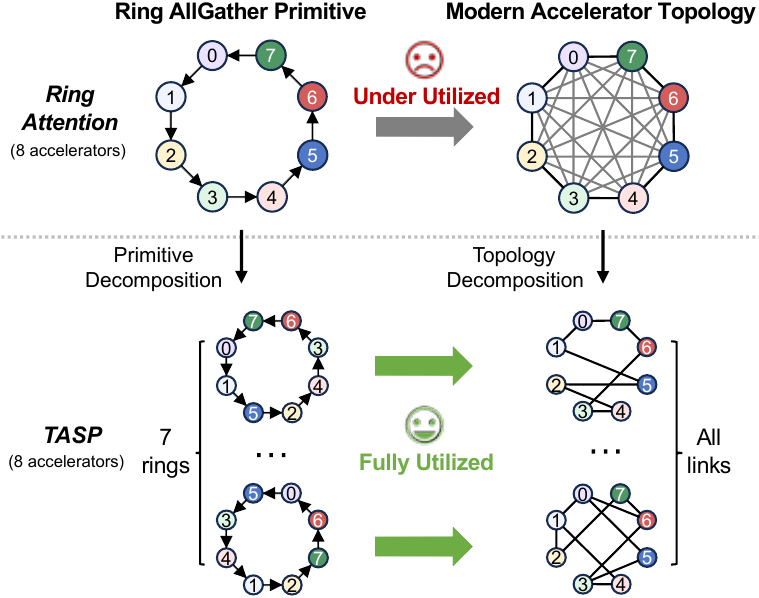}
  \caption{The mismatch between the ring-style data transfer of Ring AllGather and the fully-connected communication links of modern accelerators.}
  \label{fig:intro}
\end{figure}

\section{Introduction}

The rapid advancement of large language models (LLMs) has ushered in a new era of long-context capabilities, with modern systems now supporting context windows extending to several million tokens \cite{liu2025comprehensive,zhang2024chain}. This breakthrough is critical for complex tasks that require reasoning over massive datasets, such as processing lengthy legal documents, performing whole-codebase analysis, or engaging in extended multi-modal dialogues. However, the core architecture of the popular Transformer model presents a fundamental scalability bottleneck: the quadratic computational complexity of the self-attention mechanism \cite{vaswani2017attention}. This poses a significant challenge for inference on these long queries, as the resource demands for computation and key-value (KV) tensor storage quickly become prohibitively expensive on a single accelerator.

To address these constraints, Sequence Parallelism (SP) methods \cite{li2021sequence,jacobs2023deepspeed,brandon2023striped,li2023distflashattn,NVIDIA2023MegatronCP} have emerged as a key technique for long-context training and inference. Through distributing the prolonged sequence dimension of long-context queries across a cluster of accelerators, they effectively shard the computational load and reduce the memory footprint on each accelerator, making long-context processing more feasible. However, these approaches suffer from notable limitations: Ulysses\cite{jacobs2023deepspeed} ties the degree of parallelism to the number of Attention KV heads, which can not scale well. The other mainstream works, Ring Attention and its variants \cite{li2021sequence,brandon2023striped,li2023distflashattn,NVIDIA2023MegatronCP}, have no scalability constraint but have communication efficiency issues.
These works first chunk the query into query chunks and distribute the these chunks to multiple accelerators, and then each accelerator computes the corresponding Q chunk and KV Chunk.
Since each Q chunk requires all of the KV Chunks, an AllGather of these KV chunks must be performed.

Unfortunately, the standard AllGather primitive has a constraint that each accelerator needs to store all the KV Chunks aggregated from other accelerators, failing to reduce the memory consumption.
Thus, Ring Attention decomposes the AllGather on $n$ accelerators into $n-1$ iterations of the ring-style data transfer (as shown in Figure~\ref{fig:Multi-Ring AllGather}). In fact, this ring-style implementation of the AllGather primitive (denoted in this paper as Ring AllGather) is one of the oldest MPICH implementations of the AllGather primitive \cite{thakur2005optimization}. 
The benefits of Ring AllGather are two-fold: (1) it incurs no data duplication at each iteration of communication and computation; and (2) the communication can be fully overlapped by computation in each iteration if the compute-to-communication ratio (CCR) is above $1.0$. 
Unfortunately, Ring AllGather suffers from the problem of low communication efficiency since each iteration of its ring-style data transfer can only utilize a small portion of communication links within the AlltoAll accelerator topology.
As shown in Figure~\ref{fig:intro}, the utilization of full communication capacity is $1/(n-1)$, which is approximately 14.3\% when $n=8$. This inefficiency in communication results in a low CCR of Ring Attention on modern accelerator systems, as shown in Table~\ref{tab:ccr}. Consequently, the communication overhead of Ring Attention fails to be overlapped with computation, thus becoming the primary bottleneck that limits its overall performance.

To this end, we propose \textbf{TASP}, a novel sequence parallelism methodology designed to overcome the communication bottleneck of Ring Attention by fully leveraging all the communication capacity in modern accelerator clusters.
Our key insight is that the mismatch between the ring-style data transfer of Ring AllGather and the fully-connected communication links in modern accelerator clusters can be solved by \textbf{topology decomposition} and \textbf{primitive decomposition}, as shown in Figure~\ref{fig:intro}.
In terms of the topology decomposition, inspired by the Hamiltonian decomposition of complete directed graphs, we identify that the modern accelerator topology can be decomposed into multiple orthogonal ring datapaths that can concurrently transfer data in a ring-style without mutual interference.
We first analyze the decomposition of the AlltoAll topology of a single-node into $n-1$ edge-disjoint Hamiltonian cycles, where vertices represent accelerators and edges represent peer-to-peer communication links.
The detailed decomposition scheme is illustrated in Figure~\ref{fig:decomposition K8}, and the formulation of this scheme is presented in Section \ref{sec: decompose topology}.
Apart from the single-node scenario, Hamiltonian decomposition schemes also exist for multi-node topologies based on IB/RoCE network switches which are commonly adopted in modern AI data-centers.

\definecolor{lightred}{rgb}{1,0.9,0.9}
\definecolor{lightgreen}{rgb}{0.9,1,0.9}
\begin{table}[t]
\centering
\caption{Evaluation of CCR and speedup of TASP over Ring Attention, measured with a batch size of 48 on 8 AMD-MI300X across different sequence lengths.}
\label{tab:ccr}
\begin{tabular}{c*{5}{c}}
\toprule
{\small SeqLen} & 10K & 20K & 40K & 50K & 100K \\
\midrule
\begin{tabular}{@{}c@{}}{\small CCR}\end{tabular} & 
\cellcolor{lightred} 0.39 & 
\cellcolor{lightred} 0.65 & 
\cellcolor{lightred} 0.80 & 
\cellcolor{lightred} 0.98 & 
\cellcolor{lightgreen} 1.17 \\
{\small Speedup} & \textbf{2.4} & \textbf{1.8} & \textbf{1.5}& \textbf{1.3} & \textbf{1.1} \\
\bottomrule
\end{tabular}
\end{table}

Building upon the topology decomposition, we need to accordingly decompose the ring-style data transfer in each iteration of Ring AllGather into multiple orthogonal ring-style data transfer (We denote them all-together as Multi-Ring AllGather in this paper).
In this way, we can achieve perfect mapping between these orthogonal ring-style data transfer and the orthogonal ring datapaths obtained from topology decomposition, as shown in Figure~\ref{fig:intro}.
We prove that Multi-Ring AllGather is equivalent to Ring AllGather in terms of distributed Attention computation through a comprehensive analysis of Ring AllGather's characteristics in accessibility and zero-copy.
Moreover, considering the widely-adopted causal mask in LLMs, we present a query chunk placement strategy to further improve the workload balance between accelerators.

Through extensive experiments, we demonstrate that TASP provides a more communication-efficient solution for sequence parallelism, achieving significant speedup over the baseline method when its CCR is below $1.0$.
For single-node scenarios with 8 accelerators, TASP achieves up to 2.31$\times$ speedup on the H100 server and 3.58$\times$ speedup on the MI300X server over Ring Attention (non-causal baseline) and Zig-zag Ring Attention (causal baseline).
For multi-node scenarios, we observe an average speedup of 1.43$\times$ on two 8-H100 nodes, and an average speedup of 1.27$\times$ on four 8-H100 nodes, respectively.

Overall, this paper makes the following contributions:
\begin{itemize}
    \item We first introduce a novel methodology to improve communication efficiency via topology decomposition and primitive decomposition.

    \item We successfully apply the methodology in sequence parallelism optimization and propose TASP, inspired by Hamiltonian decomposition of complete graphs. It enables a perfect mapping between all the data transfer and all the communication links.

    \item The evaluation results demonstrate that TASP has significant performance improvement over the state-of-the-art baselines, Ring Attention and Zig-zag Ring Attention. 
\end{itemize}

\section{Background and Motivation}
\label{sec:background and motivation}

\subsection{Sequence Parallelism}

The trend toward handling long-context LLM application introduces significant challenges, including computational overhead and memory demands for KV tensor storage across extended sequences. 
To address these issues, SP has emerged as a promising technique to distribute the workload efficiently across multiple accelerators, which can be broadly categorized into two types, including Ulysses \cite{jacobs2023deepspeed} and Ring Attention \cite{liu1889ring}.

Ulysses initially shards along the sequence length dimension and distributes across accelerators, the KV tensors are partitioned by the KV head dimension through an AlltoAll operation. This allows each accelerator to maintain a complete view of the whole KV tensors along the sequence length dimension for its assigned heads. 
Since each KV head computes Attention independently, each accelerator can calculate the full Attention output for its respective heads. 
Subsequently, another AlltoAll operation can be utilized to transpose the partial Attention outputs across accelerators to form the final result. However, Ulysses's leverage of the independence between KV heads introduces a constraint: its performance is highly dependent on the number of KV heads and the computational balance across them. Current trends in Attention mechanisms exhibit fewer KV heads, such as Grouped Query Attention (GQA) with typically 4 or 8 KV heads \cite{chen2025cost}, or Multi-Query Attention (MQA) \cite{ainslie2023gqa} and Multi-Layer Attention (MLA) \cite{wang2025multi} with only a single KV head. This reduction limits Ulysses's scalability in distributed environments, as it heavily relies on a sufficient number of KV heads for effective partitioning across accelerators.
Moreover, it is hard for Ulysses to overlap its communication with computation since they are dependent to each other. 

Ring Attention partitions queries (Q and KV) along the sequence length dimension across accelerators. The Q tensors remain stationary on their respective accelerators, while the KV blocks are rotated among all accelerators through a Ring AllGather communication primitive. Over iterations, each accelerator incrementally computes, via Flash Attention \cite{dao2022flashattention}, the full Attention output for its local query segment by processing all global KV blocks. A fundamental challenge inherent to the original Ring Attention is computational imbalance. The computational cost of Attention is often unevenly distributed across the sequence length dimension, particularly under the widely used causal mask. Several optimizations have been proposed to address this issue. For instance, Striped Attention \cite{brandon2023striped} employs a striped partitioning strategy to achieve better load balance. Megatron CP \cite{NVIDIA2023MegatronCP} uses an axial-symmetric partitioning scheme that can achieve perfect load balance for causal masking. Ring Attention and its causal-masking balanced variants typically employ computation-communication overlap to mitigate the critical communication overhead. However, perfect overlapping of communication overhead is not easy to achieve. As evidenced by Table \ref{tab:ccr}, the CCR which indicates the ratio between computation and communication falls below $1.0$ under many cases. Consequently, the overall Attention performance remains constrained by the communication overhead of the Ring AllGather primitive. Therefore, improving the efficiency of communication is crucial for enhancing the performance of Ring Attention generally.

\subsection{Inter-connectivity of Hardware Platforms}

With the increasing scale of LLMs, a single accelerator can no longer meet the requirements for efficient serving. As a result, deploying LLM inference services increasingly relies on clusters of accelerators utilizing various parallelization strategies such as Tensor Parallelism (TP) and Expert Parallelism (EP) \cite{jiang2024mixtral}. These parallelization strategies, however, impose stringent demands on the inter-connectivity between devices: any pair of devices must support high-bandwidth, low-latency P2P communication, and multiple such communications should be able to occur concurrently without interference, so that collective communication operations like AlltoAll and ALLGather that are utilized in TP and EP can be efficiently carried out.

To support such capabilities, fully connected topologies have become the preferred choice for networking state-of-the-art AI accelerators, which fall into two categories: full-mesh and switch-based. For example, Huawei’s Ascend 910B NPUs \cite{liao2021ascend}, AMD’s MI300X series GPUs \cite{smith2025amd}, and Intel’s Gaudi 3 AI accelerators \cite{kaplan2024intel} all employ a full-mesh topology among 8 devices. For full-mesh topologies, vanilla Ring Attention can only utilize $\frac{1}{n-1}$ of the $n(n-1)$ unidirectional communication links among accelerators. In contrast, NVIDIA's GPUs utilize a switch-based fully connected topology \cite{choquette20213}, which allows for more flexible bandwidth allocation compared to the full-mesh approach. For larger-scale clusters, multi-level switch-based architectures—such as the CloudMatrix384 \cite{zuo2025serving}—provide fully P2P inter-connectivity among up to 384 Ascend 910 NPUs and 192 CPUs. 

These advancements of accelerator inter-connectivity underscore that high-performance P2P communication capability is both a fundamental requirement and a key trend for next-generation AI hardware architecture. However, some communication primitives can not fully utilze the communication links in an AlltoAll topology, resulting in poor communication efficiency and consequently increases end-to-end latency. Therefore, when designing and implementing parallelization strategies, it is essential to fully account for AlltoAll inter-connectivity of hardware platforms to maximize communication efficiency.

\section{Design of TASP}
The design of TASP involves two aspects: the accelerator \texttt{topology decomposition} to unveil the potential of full communication link utilization, and the Ring AllGather \texttt{primitive decomposition} to map the chunked ring-styled data transfer workload onto all the communication links. 

First, we demonstrate that modern accelerator topologies, including the single-node and multi-node ones, can be decomposed into a set of mutually orthogonal ring datapaths by Hamiltonian Decomposition. Those ring datapaths can be utilized simultaneously without interference. 

Second, we demonstrate that the \texttt{Ring AllGather} primitive utilized in Ring Attention can be decomposed into \texttt{Multi-Ring AllGather}, within which a number of ring-styled data transfer are concurrently carried out at every iteration. Consequently, we can map each ring-styled data transfer to each ring datapath, fully utilizing all communication links.

\begin{table}[t]
  \caption{Notation description.}\label{tab:symbol}
  \resizebox{0.48\textwidth}{!}{
  \begin{tabular}{l|l}
    \toprule
    Notation & Description \\
    \midrule
    $n$ & Number of accelerators \\
    $V$ & Accelerator set \\
    $E$ & Communication link set \\
    $R_i$ & The $i$-th accelerator (rank $i$) \\
    $G$ & Directed graph expression of accelerator topology \\
    $K_n$ & Complete graph with degree $n$ \\
    \bottomrule
  \end{tabular}
  }
\end{table}

\subsection{Decomposition of Accelerator Topology}
\label{sec: decompose topology}

\subsubsection{Ring Datapath Definition}
We define a ring datapath based on the formulation of the accelerator topology as a directed graph. Consider a set of accelerators denoted as $V = \{R_0, R_1, \dots, R_{n-1}\}$, where $R_i$ denotes the $i$-th rank in the communication group. Each accelerator $R_i$ has several communication ports that can be connected to other accelerators via electrical or fiber-optic cables and optionally a set of network switches. Due to duplex communication mode, a cable $c_k$ connecting $R_i$ and $R_j$ establishes two logical unidirectional communication links, denoted as $(R_i, R_j,c_k)$ and $(R_j, R_i,c_k)$. For accelerators interconnected via switches, there exist two communication links between each pair of accelerators but these links share the same amount of aggregated bandwidth.

By treating the accelerators as vertices, the set of all unidirectional communication links forms a directed edge set {by
\begin{equation}
\begin{aligned}
E = \{(R_i, R_j, c_k) \;\mid \;&\text{there exists a } \text{communication} \\
& \text{link from } R_i \text{ to } R_j \text{ via cable } c_k\}.
\end{aligned}
\end{equation}
Thus, the topology can be represented as a directed graph $G = (V, E)$.

Next, we define a ring datapath. A ring datapath on $G$ is a subgraph $D \subseteq G$ that forms a directed Hamiltonian cycle, \texttt{i.e.}, a directed cycle that visits every vertex in $V$ exactly once. Two ring datapaths $D_i = (V, E_i), D_j = (V, E_j), i \neq j$ are said to be \textbf{orthogonal ring datapaths} if they are edge-disjoint, that is,
\begin{equation}
E_i \cap E_j = \emptyset.
\end{equation}
The property ensures that a set of mutually orthogonal ring datapaths can support simultaneous data transfer without mutual interference. 

\subsubsection{Single-Node topology Decomposition}

The most common topology in modern single-node accelerators is the AlltoAll topology, encompassing switch-based (NVIDIA GPUs) and full-mesh (AMD GPUs) ones. 
Theoretically, both topologies can be represented as a complete directed graph $K_n$ of degree $n$. 
For decomposition of these AlltoAll topologies, we can directly apply the existing \emph{Hamiltonian decomposition of complete directed graphs into edge-disjoint Hamiltonian Cycles}~\cite{bae2003edge, CADA200445cubic, SAENGCHAMPA2024114197hyper, NG1997279multipartite, LIU2003305cayley} (denoted in this paper as $K_n$-decomposition).

As illustrated in Figure \ref{fig:decomposition K8}, we use 8 accelerators within a single node (\texttt{i.e.}, $K_8$) as an example, the decomposition scheme is computed via an algorithm demonstrated in Algorithm \ref{alg:hamilton1} proposed by \cite{TILLSON198068}, which decomposes a complete graph $K_n$ within $O(n)$ time complexity. 
Based on this decomposition, the $n\times (n-1)$ communication links are divided into a set of $n-1$ edge-disjoint Hamiltonian cycles. These cycles are then utilized as the mutually orthogonal ring datapaths for concurrent data transfer.

\begin{algorithm}[t]
\caption{Decomposition of $K_n$ into a set of Hamiltonian Cycles}
\label{alg:hamilton1}
\begin{algorithmic}[1]
\REQUIRE Integer $n$ with $n \ge 8$ and $n \% 4 = 0$
\ENSURE Hamiltonian cycle list $h\_cycles$
\STATE // Generate Hamiltonian paths
\STATE $init\_paths \gets \textsc{GetPath}(n-2)$
\STATE // Link the paths to form cycles  
\STATE $init\_cycles \gets \textsc{GenCycles}(init\_paths)$
\STATE // Find a set of edges in each cycle that forms a Hamiltonian Path
\STATE $k \gets n/4 - 1$
\STATE $shifts \gets \{0:1,\, k+1:4k+2,\, 2k+2:3,\, 3k+2:4k\}$
\STATE $shift \gets shifts.get(i, 2k)$
\STATE $rotated[i][j] \gets init\_cycles[i][(j + shift) \% (n-1)]$
\STATE // Disconnect these edges to form paths
\STATE $appender[rotated[i][-1]] \gets rotated[i][0]$
\STATE $appender\_list \gets [n-2]$, then append $appender[last]$ for $n-2$ steps
\STATE // Link the paths to form cycles again
\STATE $h\_cycles \gets \textsc{GenCycles}(appender)$
\RETURN $h\_cycles$
\end{algorithmic}
\end{algorithm}

\begin{figure*}
\centering
\includegraphics[width=0.95\linewidth]{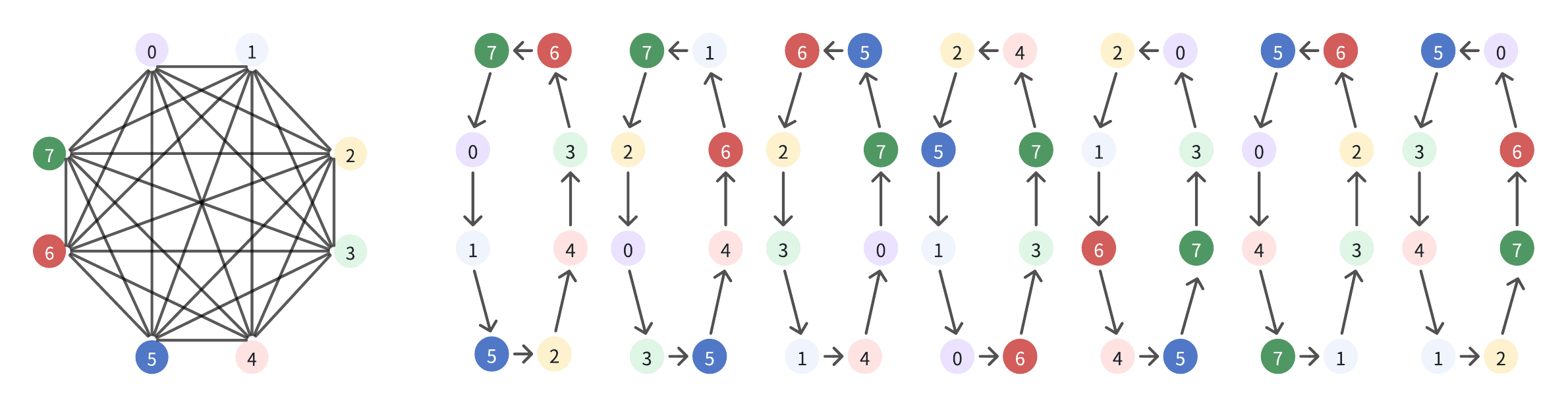}
\caption{Decomposition of 8-accelerator AlltoAll topology graph $K_8$ into 7 edge-disjoint directed Hamiltonian cycles. The decomposed Hamiltonian cycles correspond to mutually orthogonal ring datapaths that traverse all 8 accelerators.}
\label{fig:decomposition K8}
\end{figure*}

\begin{figure*}
\centering
\includegraphics[width=0.98\linewidth]{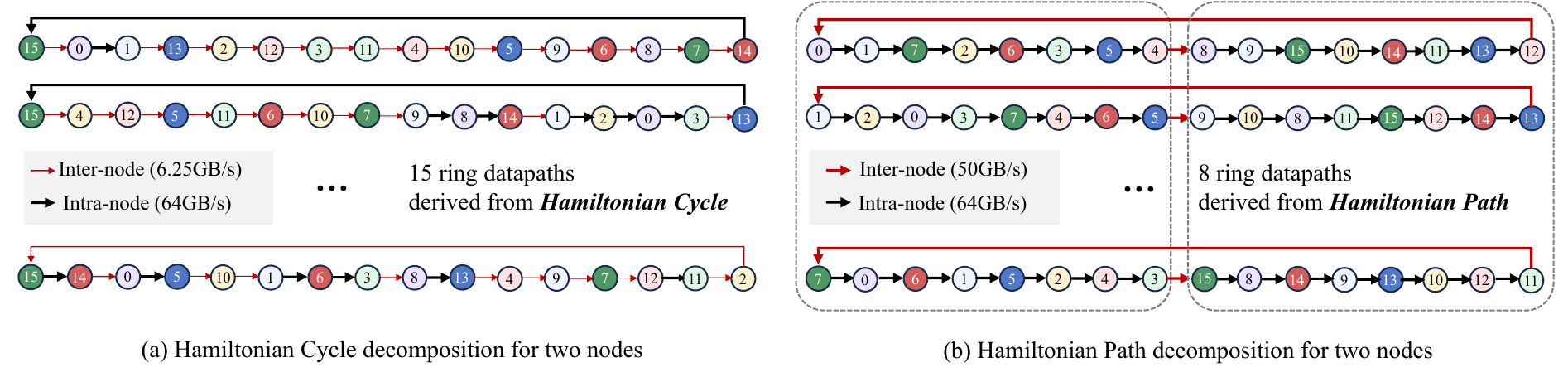}
\caption{Left: The 15 Hamiltonian cycles decomposed from the topology of H100-2 via a $K_{16}-decomposition$ scheme. Right: The 8 Hamilton cycles derived from a $(8-K_8-8)^2$-decomposition scheme.}
\label{fig:decomposition DGX H100}
\end{figure*}

\subsubsection{Multi-Node topology Decomposition}

In multi-node systems such as the NVIDIA DGX H100 GPU cluster \cite{9895480NVSWITCH}, multiple H100 servers each equipped with 8 H100 accelerators are inter-connected via a high-bandwidth Infiniband Network. Within each node, 8 accelerators are fully-connected via four ultra-high bandwidth NVSwitches, while at the same time, each accelerator is also connected with an IB NIC, making it possible for all 8 accelerators to independently and concurrently RDMA data over their own dedicated 400 Gb/s NIC.

To analyze the decomposition of such IB-connected multi-node accelerators, we first take a 2-node H100 cluster as an example (abbreviated as H100-2).
Since every two accelerators in H100-2 are either interconnected via ultra-high 7.2 TB/s (duplex) aggregated bandwidth NVLINK-based communication links or 800 GB/s (duplex) aggregated bandwidth IB-based communication links, we can simply model its topology as $K_{8\times2}$ and use the above mentioned $K_n-decomposition$ scheme for its topology decomposition, as illustrated in the left part of Figure \ref{fig:decomposition DGX H100}.  

However, this decomposition scheme will result in under-utilization of the bandwidth of intra-node NVLINK-based communication links. In a $K_{8\times2}-decomposition$ scheme there are $7\times8=56$ communication links within each node whereas $8\times 8=64$ communication links from one node to the other. Consequently, each intra-node communication link will have a bandwidth of $7200/(56*2)\approx 64 GB/s$, whereas each inter-node communication link will have a bandwidth of $\frac{800}{64*2}\approx6.3$ GB/s, an order of magnitude lower than the intra-node bandwidth. Consequently, the utilization of intra-node bandwidth is bounded by inter-node bandwidth.

To this end, we propose to model the topology of H100-2 as two $K_8$ subgraphs inter-connected with 8 bidirectional or 16 unidirectional IB-based communication links (denoted as $(m-K_m-m)^u-decomposition$ where in this case $m=8$ and $u=2$). This decomposition scheme is based on the Hamiltonian decomposition of each node into a set of \texttt{Hamiltonian paths}. A Hamiltonian path is a path in a directed graph that visits each vertex exactly once.
As illustrated in the right part of Figure \ref{fig:decomposition DGX H100}, the theoretical bandwidth of each unidirectional intra-node communication link (gray) are roughly 64 GB/s while the theoretical bandwidth of each unidirectional inter-node communication link (blue) are approximately 50 GB/s-only a 28\% difference which is significantly smaller than the over 9x difference if $K_{8\times2}$ decomposition is adopted. 

Next, we prove that the 2-node decomposition scheme can be generalized to an arbitrary $u-node$ cluster. 
If we decompose two $K_8$ subgraphs into Hamiltonian paths, we can obtain 8 Hamiltonian cycles by connecting these paths end-to-end. Since the decomposed Hamiltonian paths are at the same time a row-complete Latin Square. In each row or column of a Latin Square the entries are mutually distinct. Consequently, all 8 accelerators will appear both at the start and the end of the decomposed Hamiltonian paths, utilizing all the $2\times 8= 16$ IB NICs. Moreover, through mathematical induction, it is provable that the above decomposition scheme can scale up to an arbitrary number of H100 nodes inter-connected via IB switches. Suppose we have already decomposed a $n$-node cluster using the scheme, we can derive a decomposition of $n+1$-node cluster simply by two steps (ranks denote accelerators): (1) dis-connect the communication links from $rank\;(8n-8)\sim (8n-1)$ to $rank\; 0\sim 7$; (2) connect $rank\;(8n-8)\sim (8n-1)$ to $rank\;(8n)\sim (8n+7)$ and connect $rank\;(8n)\sim (8n+7)$ to $rank\; 0\sim 7$.

Other topologies, such as hypercubes, can also be decomposed into orthogonal ring data-paths via Hamiltonian decomposition. However, due to space limitations and the growing architectural emphasis on AlltoAll connectivity, this paper focuses primarily on the single-node and multi-node AlltoAll topologies analyzed in this section.

\begin{figure*}[t]
\centering
\includegraphics[width=0.95\textwidth]{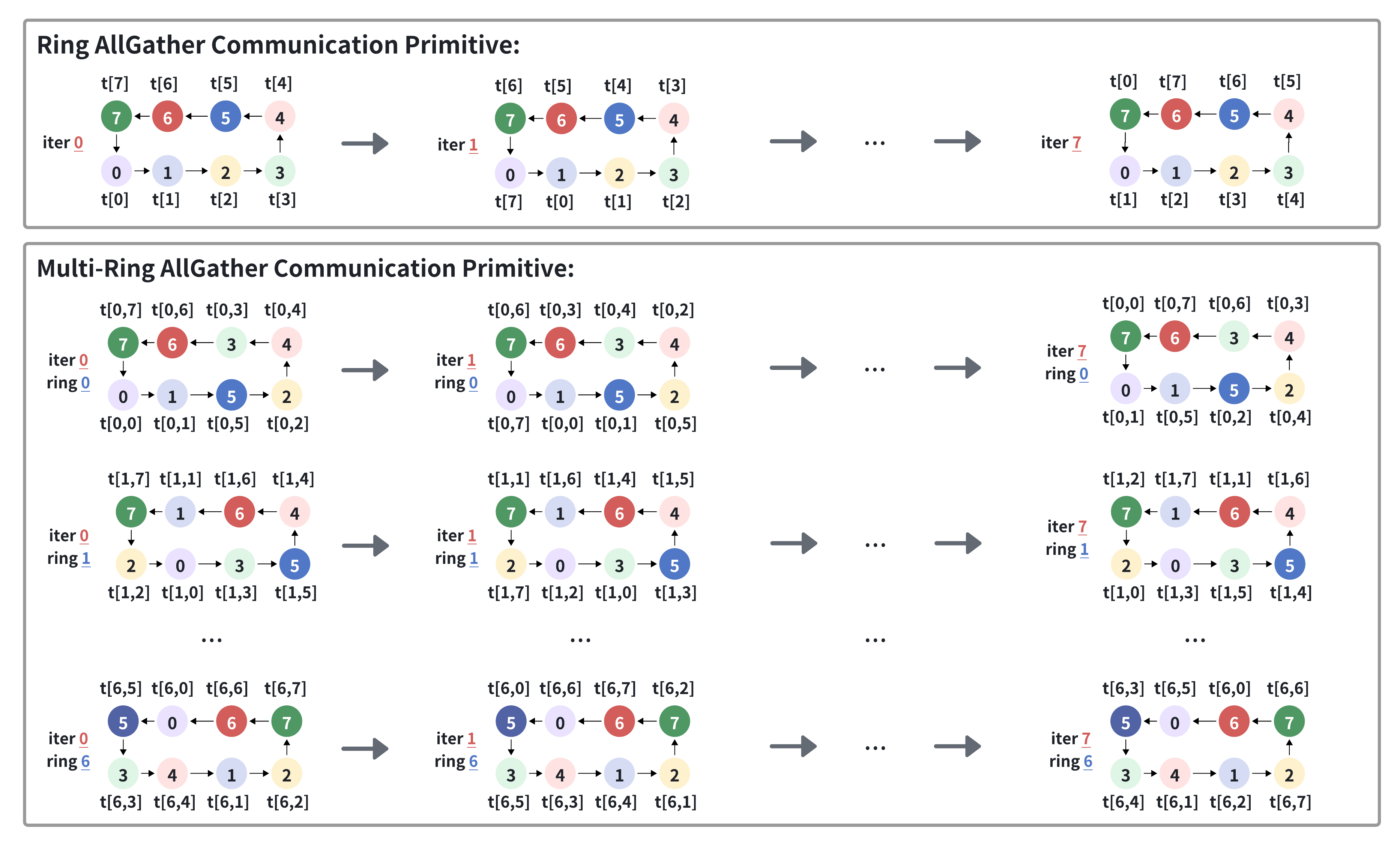}
\caption{Top: The Ring AllGather communication primitive used in Ring Attention and its variants. Colored circles represent accelerators 0–7. Symbols $t[0] \sim t[7]$ denote the transferred KV blocks. Bottom: The Multi-Ring AllGather communication primitive designed for $K_8-decomposition$. Each column illustrates one iteration with seven ring-styled data transfer for seven chunks of KV blocks. Symbol $t[i,j]$ represents the KV initially assigned at accelerator $j$ and circulates across all accelerators following the $i$-th ring-styled data transfer. Iterations 2–6 and ring-styled transfer 2-5 are omitted for brevity.}
\label{fig:Multi-Ring AllGather}
\end{figure*}

\subsection{Decomposition of Ring AllGather}
\label{sec: Primitive Decomposition}

\subsubsection{Problem Definition.}
For simplicity, we assume that for each query chunk, its Q tensors keep stationary and its KV tensors are sent to or received from other accelerators in a ring-style data transfer.
Based on the above assumption, the formulation of vanilla Ring Attention is composed of two separate components: a \texttt{Ring AllGather Communication Primitive} that assigns a query chunk at each accelerator and one ring-styled data transfer pattern for sending and receiving these chunks, as well as a \texttt{Chunk Placement Strategy} which defines the (initial) query chunk placement strategy on the accelerators.
Thus, given a Ring AllGather primitive and a Chunk Placement Strategy, the communication and computation of the vanilla Ring Attention scheme are well formulated and fully determined for every iteration.
As illustrated in Figure \ref{fig:Multi-Ring AllGather}, vanilla Ring Attention has three desirable properties.
\begin{itemize}
\item \textbf{Accessibility}: After $n-1$ iterations of KV transfer and $n$ iterations of Attention computation, every Q tensor can access every KV tensor (which are initially distributed among all accelerators) locally on its assigned accelerator.
\item \textbf{Zero-copy}: No additional memory is consumed due to KV tensor duplication.
\item \textbf{Load balance}: For bi-directional non-causal Attention, as long as the query chunks are distributed evenly among the accelerators, computation and memory load are balanced across all accelerators. For causal Attention, however, there are two types of workload balance methods, including Striped Attention \cite{brandon2023striped} and Zig-zag Ring Attention \cite{NVIDIA2023MegatronCP}.
\end{itemize}
The former two of the above properties correspond to the first component \texttt{Ring AllGather Communication Primitive} of the vanilla Ring Attention, ensuring the algorithmic correctness in Attention computation.
The third one corresponds to the \texttt{Chunk Placement Strategy} of the vanilla Ring Attention, and affects the workload balance of each iteration.
Then, we demonstrate that \texttt{Ring AllGather} can be decomposed into {Multi-Ring AllGather} while satisfying the first two properties (See Section \ref{subsubsec: Multi-Ring AllGather}).
Lastly, the chunk placement strategies for Ring Attention (Striped Attention and Zig-zag Ring Attention) cannot be directly ported to TASP. Thus, we present a Zig-zag chunk placement strategy for perfect load-balance (See Section \ref{subsubsec: zigzag hamilton}).

\subsubsection{Multi-Ring AllGather}
\label{subsubsec: Multi-Ring AllGather}

As shown in Figure \ref{fig:Multi-Ring AllGather}, compared to the original Ring AllGather primitive that only assigns one query chunk for each accelerator, an $m$-ring Multi-Ring AllGather communication primitive assigns $m$ query chunks for each accelerator and designates a specific ring-styled data transfer for each query chunk. 
The number of $m$ can be chosen arbitrarily, as long as the original query can be evenly split into $m$ query chunks. The ring-style data transfer for each query chunk can also be designed arbitrarily and mutually different. 

Since it is shown in the previous section that the topology of modern accelerators can be decomposed into a set of mutually orthogonal ring datapaths, we can thus design a Multi-Ring AllGather communication primitive 100\%-percent align with a topology decomposition. For example, we present a $7$-ring Multi-Ring AllGather Communication specifically designed for $K_8-decomposition$ of an 8-accelerator AlltoAll topology in Figure \ref{fig:Multi-Ring AllGather}. Full communication link utilization can be achieved if the ring-styled data transfer are 100\%-percent align with the ring datapaths.

However, to make sure this decomposition of Ring AllGather does not incur extra memory overhead or violate the algorithmic correctness of Attention calculation, we need to prove that it still adheres to the following properties: 

\textbf{Proof of Accessibility}: For any $Q_i$ tensor (assigned to accelerator $s_0$) and any $KV_j$ tensor, since $KV_j$ tensor traverses all accelerators along its assigned ring-styled data transfer $h_0$ (a Hamiltonian cycle), it will eventually reach accelerator $s_0$, allowing $Q_i$ to attend to $KV_j$ locally.

\textbf{Proof of Zero-copy}: Each KV tensor is assigned to exactly one ring-styled data transfer $h_0$ and is transferred solely along $h_0$, ensuring only one copy exists at any time.

\subsubsection{Zig-zag Chunk Placement for TASP}
\label{subsubsec: zigzag hamilton}
The vanilla Ring Attention adopts a naive chunk placement strategy: 
\begin{equation}
t[i]=KV[\frac{iS}{n},\frac{(i+1)S}{n})\;\text{(S is sequence length)}
\end{equation}
Translating into plain text, this strategy first evenly splits all the KV tensors into $n$ continuous blocks and map each block to accelerator, given $n$ accelerators.  

However, when a causal mask is used, the naive chunk placement strategy will disrupt computation load balance. If the naive chunk placement strategy is used together with causal masking, $i$ accelerators will become idle while only $n-i$ accelerators perform computation at the $i$-th iteration. In contrast, the Zig-Zag chunk placement strategy of Megatron CP ensures that all accelerators perform an equal amount of computation at each iteration, through a more complicated mapping: 
\begin{equation}
\begin{aligned}
t[i] = &KV\left[\frac{iS}{2n},\frac{(i+1)S}{2n}\right) \cup \\
     &KV\left[\frac{(2n-i-1)S}{2n},\frac{(2n-i)S}{2n}\right)
\end{aligned}  
\end{equation}

For perfectly load-balanced causal Attention computation, the Zig-zag chunk placement for Zig-zag Attention cannot be directly utilized in TASP. However, via some necessary adjustments, similar strategy can also be derived.
\texttt{Zig-zag TASP:} Each KV chunk $t[i,j]$ assigned at accelerator $j$ and circulated via ring-style data transfer $i$ is further split into two parts:
\begin{equation}
t[i,j][0]=KV[\frac{(2nj+i)S}{2n(n-1)}, \frac{(2nj+i+1)S}{2n(n-1)}) 
\end{equation}
\begin{equation}
t[i,j][1]=KV[S-\frac{(2nj+i+1)S}{2n(n-1)},S-\frac{(2nj+i)S}{2n(n-1)})
\end{equation}

The proof of the perfect load-balance of Zig-zag TASP is as follows:
Suppose for some accelerator $r_0$, at some iteration $iter_0$, it receives KV chunks $t_{KV}[i_0,j_0][0]$ and $t_{KV}[i_0,j_0][1]$ initially placed at accelerator $j_0$ via some ring data-path $i_0$ while at the same time the Q chunks $t_{Q}[i,r_0][0]$ and $t_{Q}[i,r_0][1]$ ($\forall i$) are kept at $r_0$. If $j<r_0$, then the token indices of $t_{Q}[i,r_0][0]$ and $t_{Q}[i,r_0][1]$ ($\forall i$) are larger than those of $t_{KV}[i_0,j_0][0]$ but smaller than those of $t_{KV}[i_0,j_0][1]$, making only $t_{KV}[i_0,j_0][0]$ needs to be attended, which is half of the transferred KV tensors via $i_0$ to accelerator $r_0$. On the contrary, if $j>r_0$, then the token indices of $t_{Q}[i,r_0][0]$ ($\forall i$) are smaller than both of $t_{KV}[i_0,j_0][0]$ and $t_{KV}[i_0,j_0][1]$ while those of $t_{Q}[i,r_0][1]$ ($\forall i$) are larger than both of $t_{KV}[i_0,j_0][0]$ and $t_{KV}[i_0,j_0][1]$, making only $t_{Q}[i,r_0][1]$ ($\forall i$)  needs to be attended, which is half of the query tokens. As a result, for any accelerator $r_0$, its computation load at any iteration is the same, leading to 100\% load-balance.

\section{Implementation of TASP}
The implementation of TASP leverages two key techniques: (1) the use of AlltoAll collective communication for concurrent data transmission across multiple ring datapaths, and (2) pre-computed routing tables to minimize CPU scheduling overhead.

\subsection{Pre-Computed Routing Tables}
To minimize CPU overhead during the distributed Attention computation, TASP pre-computes all routing tables before the forward pass. The \texttt{cal\_in/out\_mapping} functions as described in Algorithm \ref{alg:IN-OUT mapping} generate source and destination mappings for each send/receive based on a given topology decomposition scheme, eliminating the need for dynamic routing during the compute-intensive attention.

\begin{algorithm}[t]
\caption{Calculate In/Out Mapping}
\label{alg:IN-OUT mapping}
\begin{algorithmic}[1]
\REQUIRE Looping matrix $looping[m][n]$ \\ direction $d \in \{+1,-1\}$
\ENSURE Mapping matrix $mapping[n][n]$

\STATE Initialize $mapping$ as $n \times n$ matrix filled with $-1$

\FOR{$i = 0$ to $m-1$}
    \FOR{$j = 0$ to $n-1$}
        \STATE $u \gets looping[i][j]$
        \STATE $v \gets looping[i][(j + d) \bmod n]$
        \STATE $mapping[u][v] \gets i$
    \ENDFOR
\ENDFOR

\RETURN $mapping$
\end{algorithmic}
\end{algorithm}

\subsection{Concurrent Ring-styled Data Transfer via AlltoAll}

Unlike Ring Attention's utilization of batched \texttt{sendRecv} collective, TASP can utilize a single \texttt{AlltoAll} collective operation to simultaneously transmit KV blocks across all active ring datapaths (this optimization is for $K_{m\times u}$-decomposition scheme only). This approach fully leverages the low-level optimization for \texttt{AlltoAll} collective implemented in the Communication Collectives Libraries (NCCL and RCCL). 

The entire forward pass of TASP is illustrated in Algorithm~\ref{alg:hamilton}, where each accelerator prepares send buffers according to a pre-computed routing table and receives corresponding KV blocks from all other accelerators in each iteration.

\begin{algorithm}[t]
\caption{TASP Forward Pass} \label{alg:hamilton}
\begin{algorithmic}[1]
    \REQUIRE $q$, $k$, $v$, \textit{process\_group}, \textit{world\_size} 
    \ENSURE  \textit{out}, \textit{lse} 
    \STATE // Initialize routing tables:  \\
    $looping \gets gen\_hamilton\_circle(world\_size)$
    \STATE $\textit{out\_mapping} \gets \texttt{cal\_out\_mapping}(\textit{looping})$
    \STATE $\textit{in\_mapping} \gets \texttt{cal\_in\_mapping}(\textit{looping})$
    \STATE $\textit{para\_size} \gets \textit{world\_size} - 1$
    
    \STATE // Stack KV tensors: \\
    $\textit{this\_kv} \gets \left[ \textit{stack}(k[i], v[i]) \ \forall i \in \{0, \textit{para\_size}\} \right]$
    \STATE // Initialize receive buffers: \\
    $\textit{next\_kv} \gets \left[ \textit{empty\_like}(\textit{this\_kv}[0]) \ \forall i \in \{0, \textit{para\_size}\} \right]$

    \FOR{$\textit{step} = 0$ \TO $\textit{world\_size} - 1$}
        \STATE // Extract K, V: \\
        $k \gets \left[ \textit{this\_kv}[i][0] \right]$, $v \gets \left[ \textit{this\_kv}[i][1] \right]$
        
        \IF{$\textit{step} < \textit{world\_size} - 1$}
            \STATE // Prepare send/recv buffers using routing tables:
            \STATE $\textit{send\_buf} \gets \left[ \textit{this\_kv}[\textit{out\_mapping}[i]]\right]$          
            \STATE $\textit{recv\_buf} \gets \left[ \textit{next\_kv}[\textit{in\_mapping}[i]]\right]$
            \STATE \textit{All-to-All}(\textit{send\_buf}, \textit{recv\_buf}) 
        \ENDIF
            
        \STATE // Concatenate KV blocks: \\
        $k_{\text{cat}} \gets \textit{concat}(k, \text{dim}=\textit{seqlen})$\\
        $v_{\text{cat}} \gets \textit{concat}(v, \text{dim}=\textit{seqlen})$
        \STATE // Compute Attention: \\
        $\textit{output}, \textit{lse\_curr} \gets \texttt{flash\_attn}(q, k_{\text{cat}}, v_{\text{cat}})$ \\
        $\textit{out}, \textit{lse} \gets \texttt{update\_out\_lse}(\textit{out}, \textit{lse}, \textit{output}, \textit{lse\_curr})$
        
        \IF{$\textit{step} < \textit{world\_size} - 1$}
            \STATE // \textit{Wait} for communication completion
            \STATE $\textit{this\_kv} \gets \textit{next\_kv}$, $\textit{next\_kv} \gets \textit{this\_kv}$
        \ENDIF
    \ENDFOR
    \RETURN $\textit{out}, \textit{lse}$
\end{algorithmic}
\end{algorithm}

\section{Evaluation}

\begin{figure*}[t]
    \centering
    \begin{subfigure}[b]{0.5\textwidth} 
        \centering
        \includegraphics[width=\textwidth]{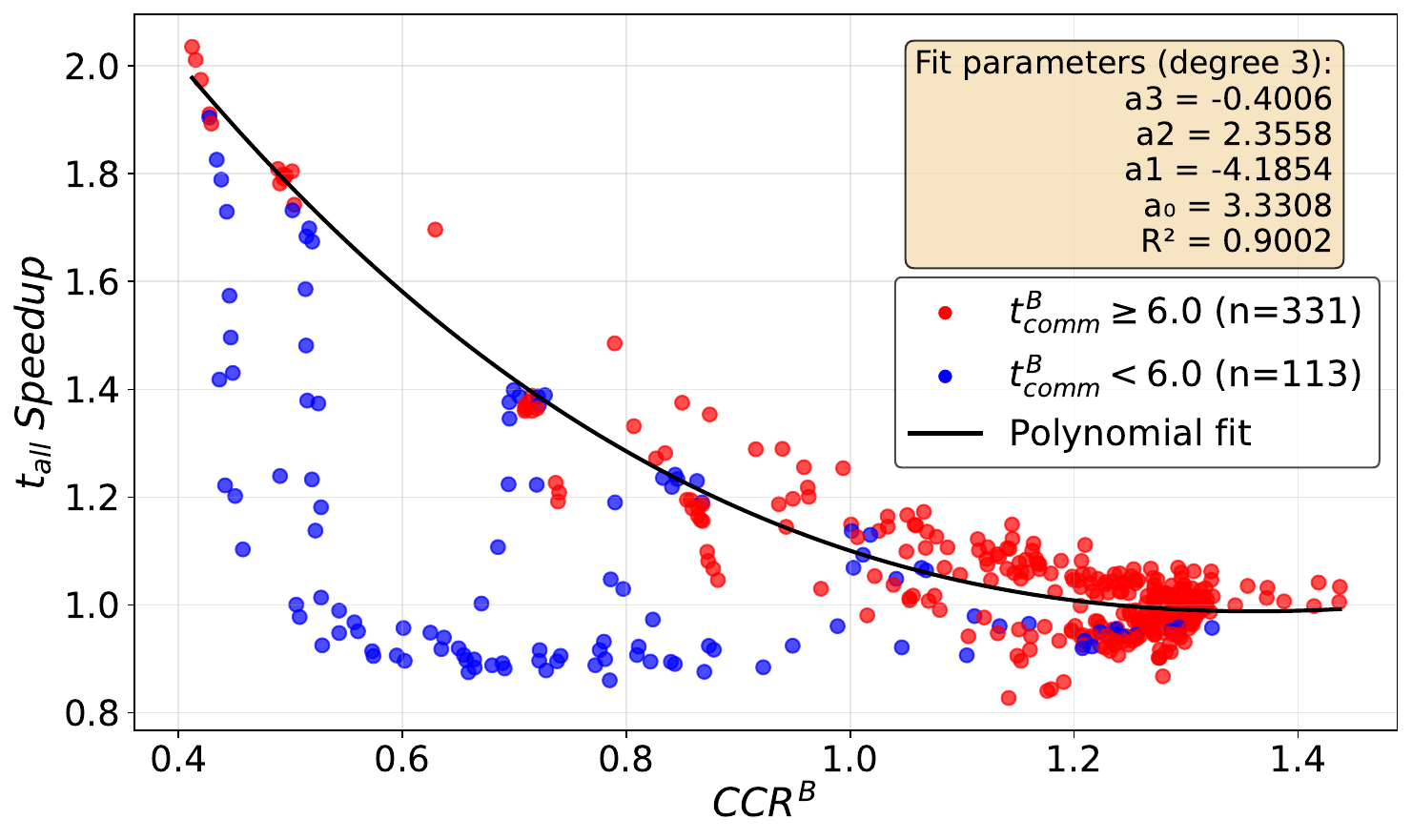}
        \caption{H100-1 (full mask)}
        \label{fig:sub1}
    \end{subfigure}
    \hfill 
    \begin{subfigure}[b]{0.49\textwidth} 
        \centering
        \includegraphics[width=\textwidth]{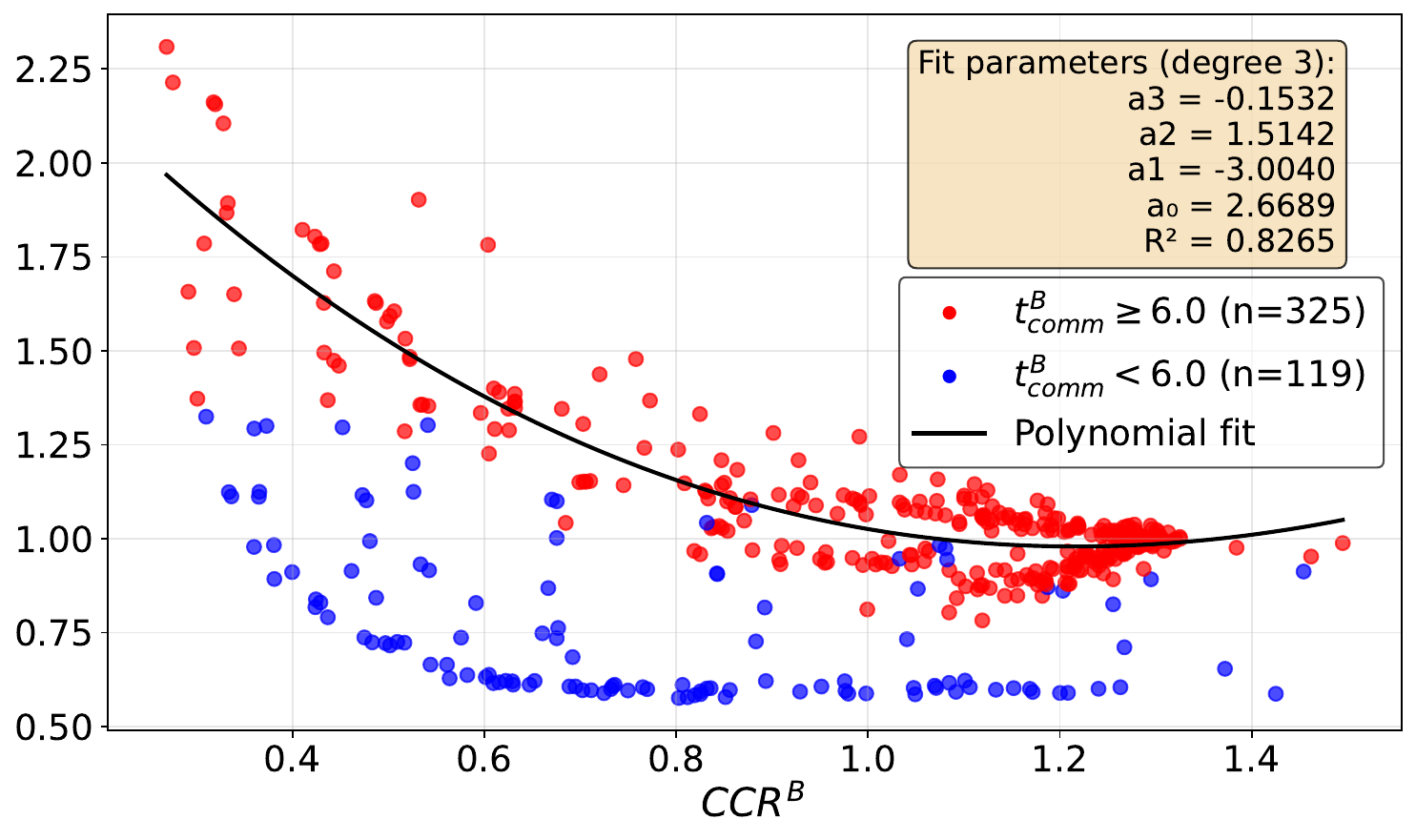}
        \caption{H100-1 (causal mask)}
        \label{fig:sub2}
    \end{subfigure}
    
    
    \begin{subfigure}[b]{0.5\textwidth}
        \centering
        \includegraphics[width=\textwidth]{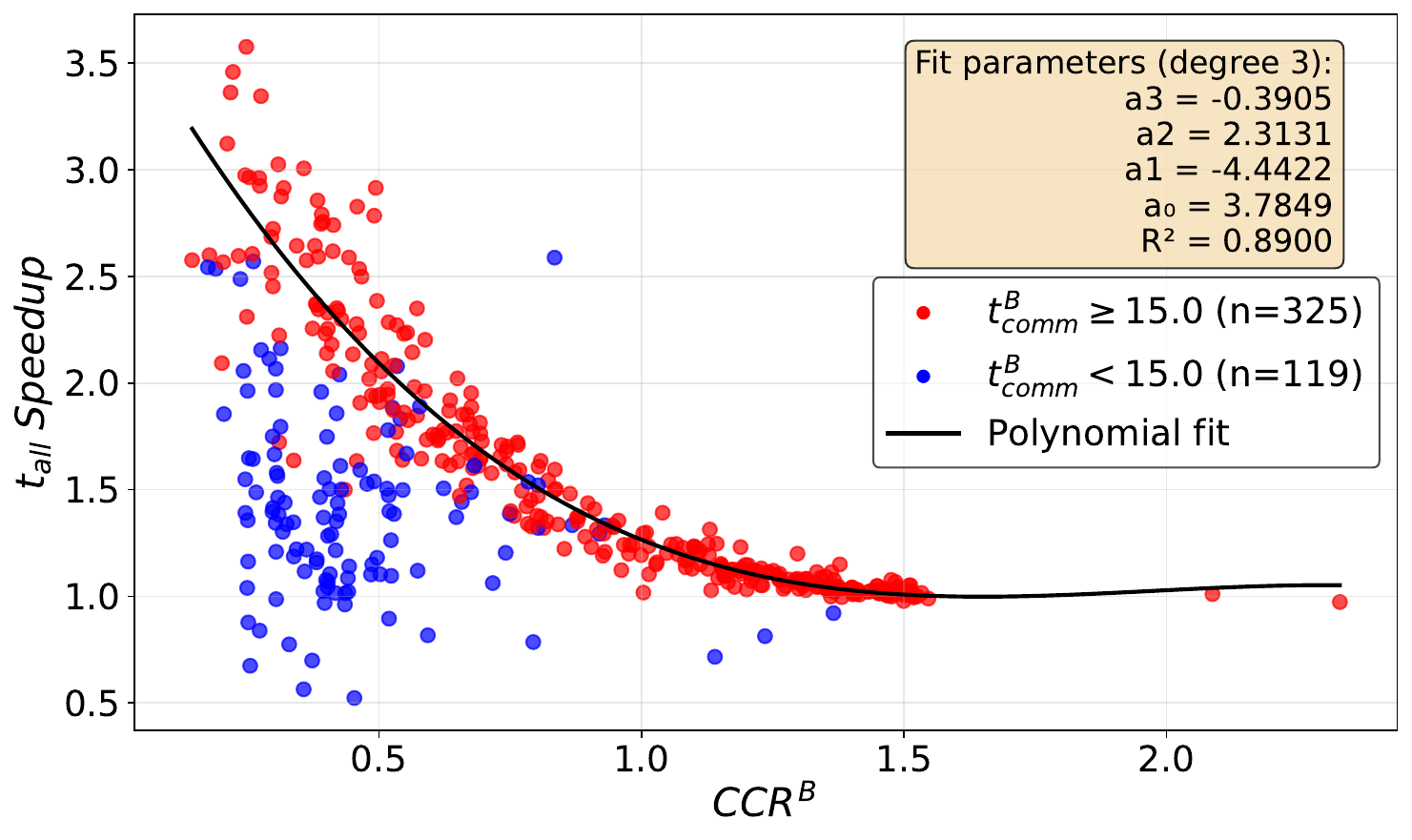}
        \caption{MI300X-1 (full mask)}
        \label{fig:sub3}
    \end{subfigure}
    \hfill
    \begin{subfigure}[b]{0.49\textwidth} 
        \centering
        \includegraphics[width=\textwidth]{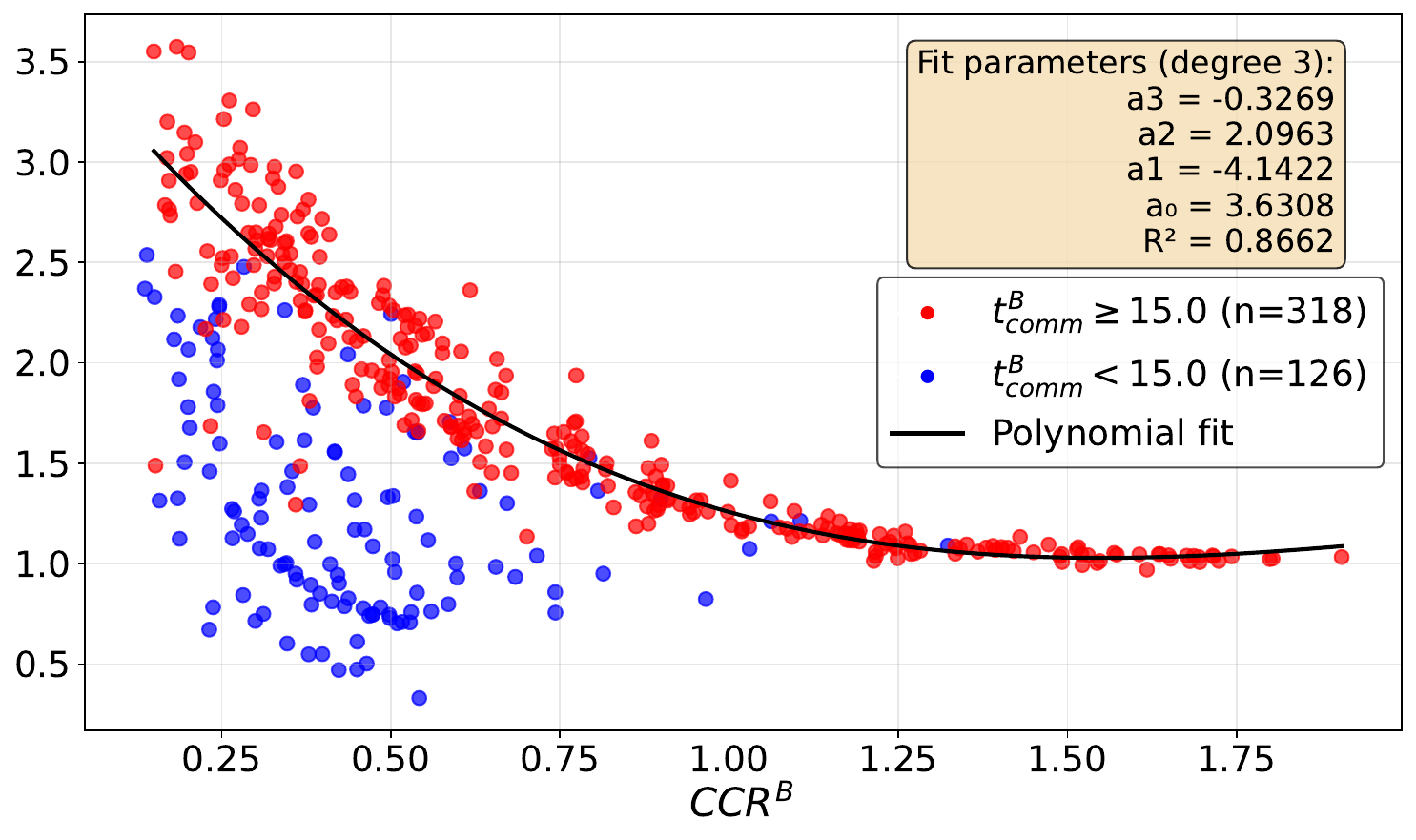}
        \caption{MI300X-1 (causal mask)}
        \label{fig:sub4}
    \end{subfigure}
    
    \caption{$t_{all}$ speedup v.s. $CCR^B$. The test cases are divided into two sets (colored red and blue) to distinguish the data transfer size impact on our communication optimization. For cases with sufficient data volume to saturate the bandwidth (dots in red), TASP achieves a significant $t_{all}$ speedup compared to baselines. Furthermore, the speedup exhibits a decreasing trend with the increase of $CCR^B$, as the optimized part lies in communication. }
    \label{fig:total-single-node}
\end{figure*}

\begin{figure*}[t]
    \centering
    \begin{subfigure}[b]{0.5\textwidth}
        \centering
        \includegraphics[width=\textwidth]{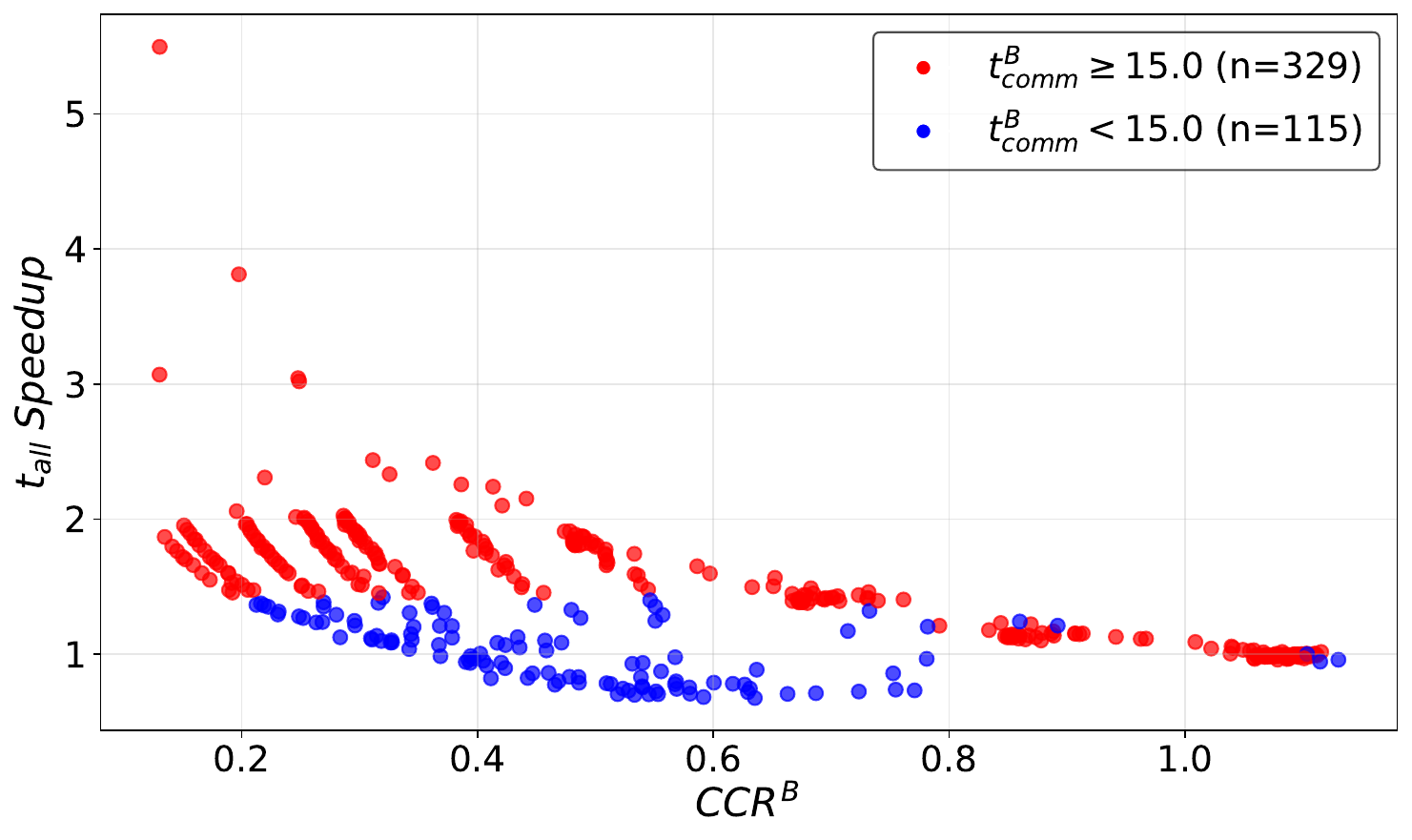}
        \caption{H100-2 ($K_{8\times2}$)}
        \label{fig:sub1}
    \end{subfigure}
    \hfill 
    \begin{subfigure}[b]{0.49\textwidth} 
        \centering
        \includegraphics[width=\textwidth]{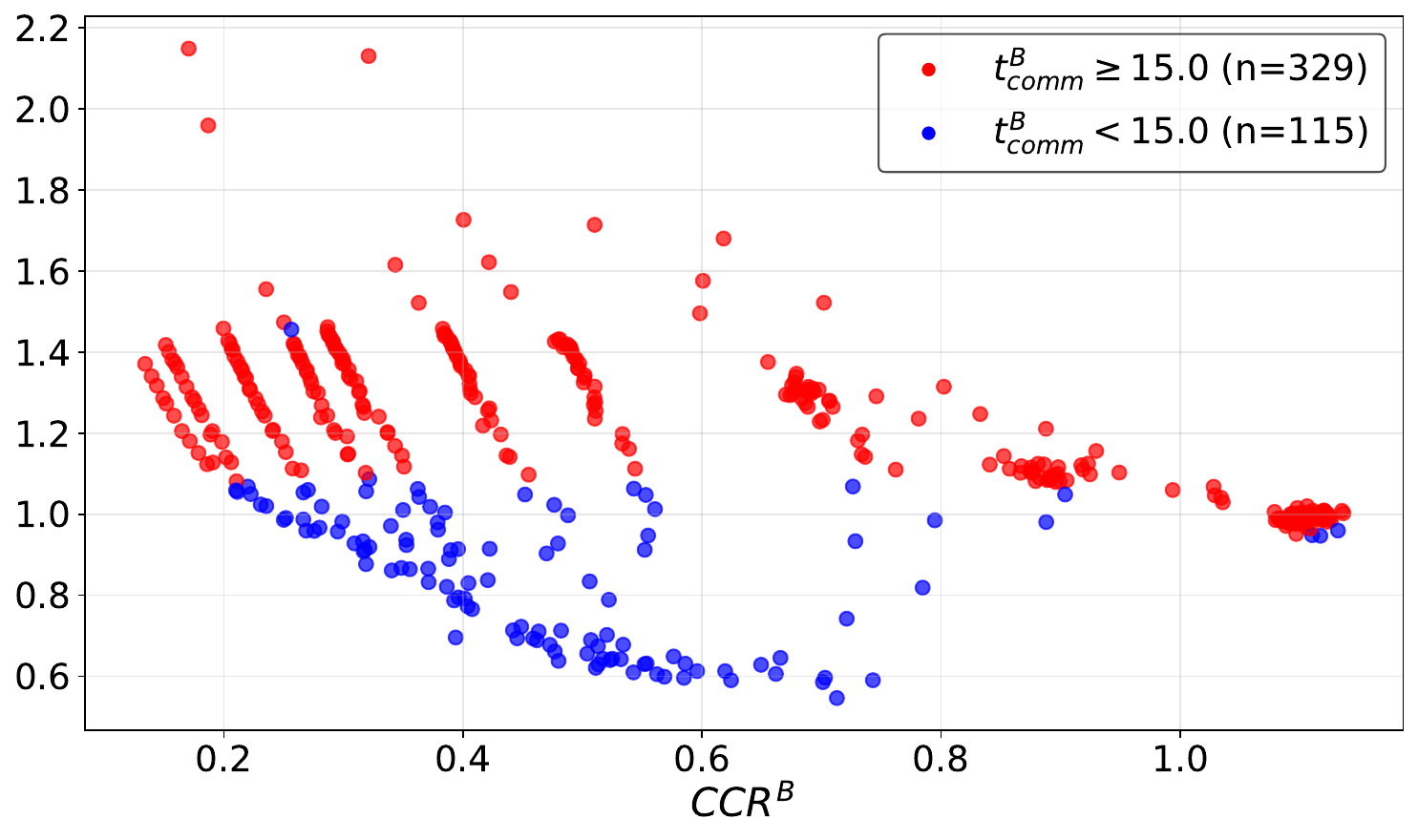}
        \caption{H100-2 ($(8-K_8-8)^2$)}
        \label{fig:sub2}
    \end{subfigure}
    
    
    \begin{subfigure}[b]{0.5\textwidth}
        \centering
        \includegraphics[width=\textwidth]{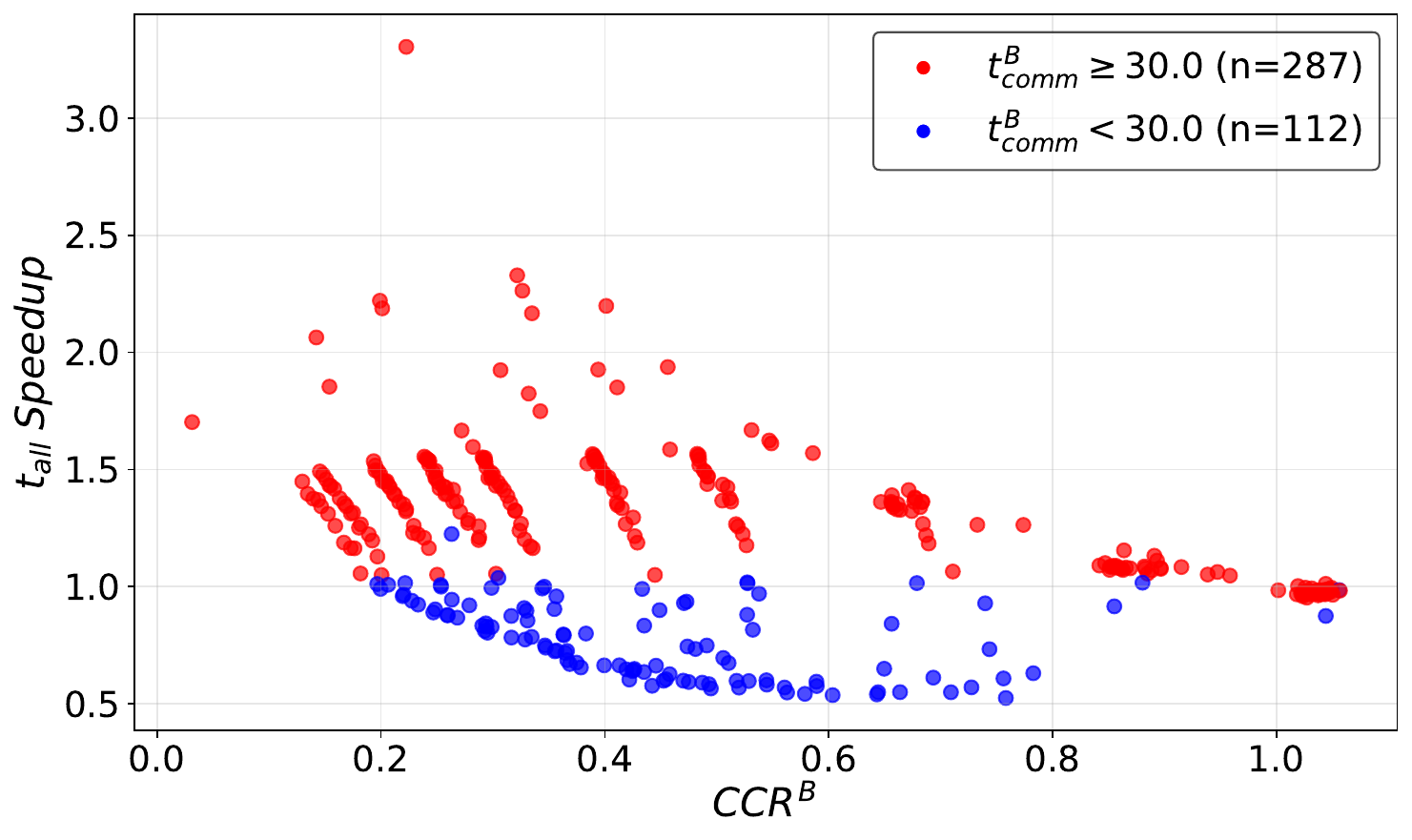}
        \caption{H100-4 ($K_{8\times4}$)}
        \label{fig:sub3}
    \end{subfigure}
    \hfill
    \begin{subfigure}[b]{0.49\textwidth} 
        \centering
        \includegraphics[width=\textwidth]{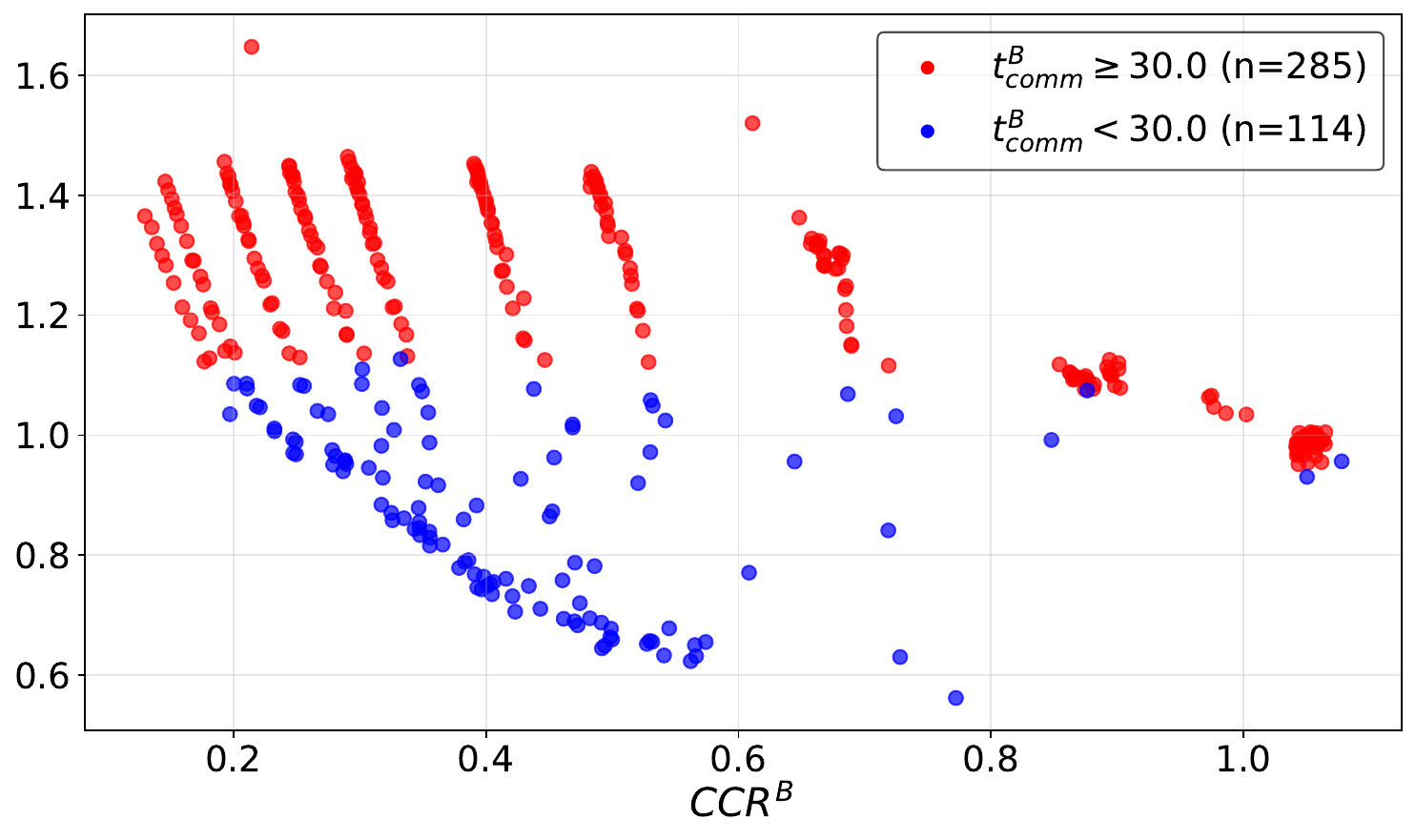}
        \caption{H100-4 ($(8-K_8-8)^4$)}
        \label{fig:sub4}
    \end{subfigure}
    
    \caption{$t_{all}$ speedup v.s. $CCR^B$. Similar to the single-node evaluation, the test cases are divided into two sets (colored blue and red) by communication volume size, and the speedup exhibits a decreasing trend with
    the increase of $CCR^B$. As the number of accelerator node scales up, TASP (with $(m-K_m-m)^u$-decomposition scheme) maintains an almost constant speedup ratio, demonstrating better scalability.}
    \label{fig:total-multi-node}
\end{figure*}

\subsection{Setup}

\subsubsection{Testbed} We evaluate TASP on four distinct hardware platforms: (1) A single node with eight NVIDIA H100 GPUs interconnected via NVSwitch4 with 3.6TB/s bisection bandwidth. (2) A single node with eight AMD MI300X GPUs in a full-mesh topology via AMD Infinity Fabric with 896 GB/s aggregated bandwidth. (3) A two-node server, each node equipped with eight NVIDIA H100 GPUs. The two nodes are interconnected via 16 NVIDIA Mellanox Infiniband NICs (eight NICs for each node) with 800 GB/s (duplex) aggregate bandwidth per node. (4) A four-node server with the same intra-node and inter-node configurations as (3).


\subsubsection{Benchmark. } We evaluate the performance of TASP on three sets of approximately 400 test cases with various input sizes (1287 test cases in total). Specifically, the batch size ranges from 1 to 128, the sequence length from 3k to 1M, and the number of Attention heads in $\{4,12,20\}$ with a head dimension of 64. We use BF16 precision for experiments, nevertheless, it is straightforward to adapt to other precisions.

For different hardware platforms, we set different maximum sequence lengths for test cases. For all single-node platforms, we use a maximum sequence length of 430K. For two-node and four-node platforms, the maximum sequence lengths are 490K and 1M, respectively.

\subsubsection{Baseline.} We use Ring Attention~\cite{liu1889ring} as our baseline method for cases with non-causal masks and Zig-zag Ring Attention~\cite{NVIDIA2023MegatronCP} for cases with causal masks. 
\begin{itemize}
    \item Ring Attention~\cite{liu1889ring}. The fundamental distributed Attention implementation specialized for sequence parallelism. 
    \item Zig-zag Ring Attention~\cite{shoeybi2019megatron}. The distributed Attention implementation modified for Attention with causal masks, to address the workload imbalance in Ring Attention. 
\end{itemize}
For the fairness of comparison, all methods (including TASP) are implemented in PyTorch~\cite{paszke2019pytorch} and utilize the same Flash Attention~\cite{dao2022flashattention} implementation. We adopt the open source code~\cite{ring-flash-attention} of Ring Attention and Zig-zag Ring Attention for baseline comparison. 

\subsubsection{Metrics. } We collect three metrics for evaluation, \textit{i.e.}, \textit{$t_{comm}$} denoting the total communication time, \textit{$t_{comp}$} denoting the total computation time and \textit{$t_{all}$} denoting the overall latency of each method. Since we implement TASP and baseline methods using two CUDA/HIP streams to overlap communication and computation, we use the execution time of each stream counted via CUDA/HIP events to derive \textit{$t_{comm}$} and \textit{$t_{comp}$}, and adding them up to derive \textit{$t_{all}$}. 

Moreover, to explore the impact of the (latency) ratio of computation and communication on the performance of our proposed method, we use $CCR^{B}$ which is the computation-to-communication ratio of the baseline method at each test case as the independent variable (x-variable) in our analysis.

\subsection{Single-node Evaluation}

We first evaluate the single-node performance of TASP, with the eight NVIDIA H100 GPU server (abbreviated as H100-1) representing the switch-based topology and the eight AMD-MI300X GPU server (abbreviated as MI300X-1) representing the full-mesh topology. 

As depicted in Figure~\ref{fig:total-single-node}, we divide the test cases into two sets to evaluate the performance of TASP. The first set includes the test cases with relatively large communication latency ($t^{B}_{comm} \geq T$, $T$ is a threshold varying with hardware platform's bandwith capacity), which can fully occupy the interconnection bandwidth of accelerators, and the second set includes the test cases with $t^{B}_{comm} < T$, failing to saturate the bandwidth. In Figure~\ref{fig:total-single-node}, the first set is depicted in red and the second set in blue.

We observe that TASP achieves significant speedups across the first set regarding the overall time $t_{all}$, yielding an average speedup of 1.05$\times$ and 1.08$\times$ for full-mask and casual-mask cases on H100-1, and an average speedup of 1.57$\times$ and 1.68$\times$ for full-mask and casual-mask cases on MI300X-1, respectively. Moreover, TASP achieves up to \textbf{2.31$\times$} maximum speedup on H100-1 and \textbf{3.58$\times$} maximum speedup on MI300X-1 over the baselines, respectively. For the communication time \textit{$t_{comm}$} alone, TASP achieves a maximum of a \textbf{4.72$\times$} speedup on MI300X-1 and \textbf{2.87$\times$} speedup on H100-1 against the baseline methods (not shown in the figure).

\textbf{\textit{The impact of $CCR^B$.}} Through evaluation, we find that a lower computation-communication ratio ($CCR^B$) enables TASP to be more communication efficient than baseline methods and yield a higher speedup in $t_{all}$.
The lower the $CCR^B$, the baseline method is more communication-bound, and the higher the $t_{all}$ speedup ratio is achieved. 

\textbf{\textit{The impact of mask pattern.}} TASP achieves a higher speedup on causal-masked test cases on both H100-1 and MI300X-1, since causal-masked Attention has theoretically two times lower $CCR^B$ than full-masked Attention. 

\textbf{\textit{The impact of hardware platform.}} Besides, we note that TASP achieves much superior performance optimization on MI300X-1 over H00-1 in most test cases. This is not only attributed to the much lower intra-node communication bandwidth of MI300X-1 that makes it more communication-bound in most test cases, but also aligns with our theoretical analysis of full-mesh topology \textit{v.s.} switch-based topology in Section \ref{sec: decompose topology}.

\subsection{Multi-node Evaluation}

The multi-node evaluation results are depicted in Figure~\ref{fig:total-multi-node}, and the two-node and four-node servers equipped with H100 GPUs are denoted as H100-2 and H100-4, respectively. To demonstrate the efficiency of the proposed topology decomposition method for multiple nodes, we compare the two distinct schemes, \textit{i.e.}, $K_{m\times u}$-decomposition and $(m-K_m-m)^u$-decomposition, where $m$ denotes the accelerator number within a node and $u$ denotes the node number. For simplicity, we only present the results of the full-masked TASP on those multi-node platforms. 

Similar to the single-node evaluation, we also present the results in two sets (red and blue) to distinguish the communication data volume impact. In the following, we focus on the test cases with sufficient communication sizes (the set in red). We observe an average speedup of 1.43$\times$ with $K_{8\times2}$-decomposition and 1.20$\times$ with $(8-K_8-8)^2$-decomposition on H100-2, and an average speedup of 1.27$\times$ with $K_{8\times4}$-decomposition and 1.20$\times$ with $(8-K_8-8)^4$-decomposition on H100-4, respectively.

According to Section~\ref{sec: decompose topology}, the $(m-K_m-m)^u$-decomposition theoretically outperforms the $K_{m\times u}$-decomposition, due to the topology-aware communication link utilization. However, the experiments yield the opposite results when there are only two nodes. This is due to the difference in the implementation of these two decomposition schemes, as the $K_{m\times u}$-decomposition scheme can directly call the AlltoAll API from NCCL/RCCL while the $(m-K_m-m)^u$-decomposition scheme necessitates utilizing the batched \textit{SendRecv} to customize the communication, which leads to communication efficiency degradation due to lack of low-level optimization. However, we observe that the $(m-K_m-m)^u$-decomposition is much more scalable than $K_{m\times u}$-decomposition, as its speedup remains almost unchanged when transferring from H100-2 to H100-4, while the speedup of $K_{m\times u}$-decomposition significantly reduces. 

\subsection{Overhead Analysis}

The overhead of TASP lies in both computation and communication. 
In terms of communication, TASP employs a complex communication primitive (the AlltoAll collective).
The batched \textit{SendRecv} employed by $(m-K_m-m)^u$-decomposition-based TASP is also more complicated than the batched \textit{SendRecv} employed in vanilla Ring Attention ($m$ vs $2$ simultaneous SendRecvs are batched).
While effective in utilizing all orthogonal ring datapaths, this incurs higher runtime overhead compared. Consequently, this overhead degrades overall performance in scenarios with smaller communication volume, such as the test cases with small batch sizes and short sequence lengths, as reflected by the blue data points in Figure \ref{fig:total-single-node} and Figure \ref{fig:total-multi-node}. However, in other scenarios (approximately 3/4 of all test cases), this overhead is negligible compared to the communication latency.

The additional computational overhead of TASP primarily stems from its fine-grained KV Cache partitioning. For instance, in a $K_8$-decomposition–based TASP scheme, the KV Cache is split into 56 chunks (7 chunks per accelerator), whereas vanilla Ring Attention requires only 8 chunks. Although the routing and mapping of these KV chunks can be precomputed, the increased chunk number introduces slightly higher CPU overhead compared to vanilla Ring Attention. Fortunately, this CPU overhead does not affect the overall execution time, as the kernel launch latency of TASP can be fully overlapped with GPU computation—owing to the absence of GPU–CPU synchronization.

Beyond CPU overhead, our current implementation of TASP also introduces minor GPU computation overhead compared to vanilla Ring Attention, as all KV Cache chunks must be concatenated on each rank before invoking the flash-attn kernel for Attention computation. Empirical measurements indicate that this additional overhead is modest, resulting in only a 1\%–5\% slowdown in the total computation time ($t_{\text{comp}}$). Nevertheless, we note that such overhead can be further minimized through kernel optimization techniques.

\section{Related Work}
\noindent\textbf{Efficient attention mechanism.} The attention mechanism in Transformer models has long suffered from low device efficiency on modern accelerators. To bridge this gap, researchers have proposed various improvements. For instance, Linformer \cite{wang2020linformer} and Sparse Transformer \cite{child2019generating} attempted to replace standard attention computation with linear approximation and sparsification methods, respectively. Rabe and Staats \cite{rabe2021self} introduced a memory-efficient self-attention algorithm that reduces memory complexity from $O(n^2)$ to $O(1)$. FlashAttention \cite{dao2022flashattention, dao2023flashattention} provided an efficient implementation by leveraging custom CUDA kernels and optimizations tailored to the GPU memory hierarchy. 

\noindent\textbf{Attention for long sequences.} As the context length of LLMs extends to the million-token level, the attention mechanism faces significant computational overhead and memory pressure. Data parallelism \cite{dean2012large}, model parallelism \cite{jia2019beyond}, pipeline parallelism \cite{huang2019gpipe}, tensor parallelism \cite{shoeybi2019megatron}, and expert parallelism \cite{singh2023hybrid}, cannot meet the requirements. To address this challenge, Li et al. \cite{li2021sequence,korthikanti2023reducing} proposed sequence parallelism. Ulysses \cite{jacobs2023deepspeed} improves the communication efficiency of SP by replacing all-gather and reduce-scatter operations with smaller AlltoAll operations. Similarly, LightSeq \cite{li2023lightseq} achieves communication-computation overlap by partitioning computational tasks into bubbles. Ring Attention \cite{liu1889ring} involves partitioning queries and the KV Cache along the sequence dimension across different accelerators, keeping queries stationary while rotating KV Cache segments through ring communication. Striped Attention \cite{brandon2023striped} addresses load imbalance by employing a striped partitioning scheme. Megatron CP \cite{NVIDIA2023MegatronCP} can achieve fully balanced loads across accelerators, but it becomes ineffective when applied to sparse attention mechanisms. In short, Ring Attention and its variants are constrained by the limited communication efficiency inherent to ring-based communication paradigms.

\section{Limitation and Discussion}

Our proposed algorithm's effectiveness in long-sequence multi-accelerator inference is closely tied to the CCR: when the ratio is low, our communication-focused optimizations target the primary bottleneck (communication) to deliver notable performance gains; yet in compute-intensive attention scenarios—especially with a CCR exceeding 1.5—such gains vanish.

This does not indicate flaws in the algorithm itself, but rather that the overall latency of such distributed Attention method is shaped by multiple factors. When the impact of other factors surpasses that of the communication overhead, the effectiveness of communication-oriented optimizations will naturally diminish, which clarifies the algorithm’s strengths in collaborating with computation-oriented optimizations such as sparse Attention.
\section{Conclusion}
We present TASP, a novel SP method that leverages the Hamiltonian decomposition of complete directed graphs to optimize the attention prefill phase for long-context LLMs. To overcome the limitations of existing methods, TASP employs two steps of optimization: topology decomposition and primitive decomposition to fully utilize the native interconnect topology of AI accelerators. Through experimental evaluation, TASP achieves higher communication efficiency than Ring Attention (and Zig-zag Ring Attention) in both single-node and multi-node topologies, underscoring the critical importance of topology-aware optimization in designing high-performance SP methods.

\bibliographystyle{unsrt}  
\bibliography{references}

\end{document}